\documentclass[conference]{IEEEtran}
\IEEEoverridecommandlockouts
\usepackage{cite}
\usepackage{amsmath,amssymb,amsfonts}
\usepackage{algorithmic}
\usepackage{graphicx}
\usepackage{textcomp}
\usepackage{xcolor}

\usepackage{bbding}
\usepackage{tabularx}
\usepackage{multirow}
\usepackage{makecell}
\usepackage{booktabs}
\usepackage{subfigure}

\def\BibTeX{{\rm B\kern-.05em{\sc i\kern-.025em b}\kern-.08em
    T\kern-.1667em\lower.7ex\hbox{E}\kern-.125emX}}

\makeatletter
\newcommand{\linebreakand}{%
  \end{@IEEEauthorhalign}
  \hfill\mbox{}\par
  \mbox{}\hfill\begin{@IEEEauthorhalign}
}

\begin{document}

\title{RNG: Reducing Multi-level Noise and Multi-grained Semantic Gap for Joint Multimodal Aspect-Sentiment Analysis
}

\author{
\IEEEauthorblockN{Yaxin Liu}
\IEEEauthorblockA{\textit{Institute of Information Engineering,} \\
\textit{Chinese Academy of Sciences}\\
\textit{School of Cyber Security,} \\
\textit{UCAS}\\
Beijing, China \\
liuyaxin@iie.ac.cn}
\and
\IEEEauthorblockN{Yan Zhou\textsuperscript{*}\thanks{\textsuperscript{*}Corresponding author.}}
\IEEEauthorblockA{\textit{Institute of Information Engineering,} \\
\textit{Chinese Academy of Sciences}\\
\textit{School of Cyber Security,} \\
\textit{UCAS}\\
Beijing, China \\
zhouyan@iie.ac.cn}
\and
\IEEEauthorblockN{Ziming Li}
\IEEEauthorblockA{\textit{Institute of Information Engineering,} \\
\textit{Chinese Academy of Sciences}\\
\textit{School of Cyber Security,} \\
\textit{UCAS}\\
Beijing, China \\
liziming@iie.ac.cn}
\linebreakand
\IEEEauthorblockN{Jinchuan Zhang}
\IEEEauthorblockA{\textit{Institute of Information Engineering,} \\
\textit{Chinese Academy of Sciences}\\
\textit{School of Cyber Security,} \\
\textit{UCAS}\\
Beijing, China \\
zhangjinchuan@iie.ac.cn}
\and
\IEEEauthorblockN{Yu Shang}
\IEEEauthorblockA{\textit{Academy of Cyber} \\
Beijing, China \\
18600696279@163.com}
\and
\IEEEauthorblockN{Chenyang Zhang}
\IEEEauthorblockA{\textit{Academy of Cyber} \\
Beijing, China \\
chenyangsat@gmail.com}
\and
\IEEEauthorblockN{Songlin Hu}
\IEEEauthorblockA{\textit{Institute of Information Engineering,} \\
\textit{Chinese Academy of Sciences}\\
\textit{School of Cyber Security,} \\
\textit{UCAS}\\
Beijing, China \\
husonglin@iie.ac.cn}
}

\maketitle

\begin{abstract}
As an important multimodal sentiment analysis task, Joint Multimodal Aspect-Sentiment Analysis (JMASA), aiming to jointly extract aspect terms and their associated sentiment polarities from the given text-image pairs, has gained increasing concerns. Existing works encounter two limitations: (1) multi-level modality noise, i.e., instance- and feature-level noise; and (2) multi-grained semantic gap, i.e., coarse- and fine-grained gap. Both issues may interfere with accurate identification of aspect-sentiment pairs. To address these limitations, we propose a novel framework named RNG for JMASA. Specifically, to simultaneously reduce multi-level modality noise and multi-grained semantic gap, we design three constraints: (\romannumeral1) Global Relevance Constraint (GR-Con) based on text-image similarity for instance-level noise reduction, (\romannumeral2) Information Bottleneck Constraint (IB-Con) based on the Information Bottleneck (IB) principle for feature-level noise reduction, and (\romannumeral3) Semantic Consistency Constraint (SC-Con) based on mutual information maximization in a contrastive learning way for multi-grained semantic gap reduction. Extensive experiments on two datasets validate our new state-of-the-art performance.
\end{abstract}

\section{Introduction}
\label{sec:introduction}
\begin{figure}[!t]
\centering
\includegraphics[width=3.3in]{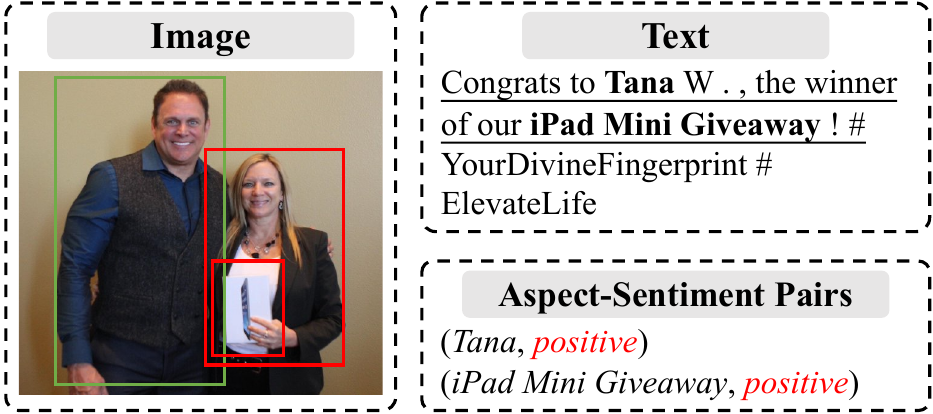}
\caption{An example of JMASA.}
\label{fig:example}
\end{figure}

Joint Multimodal Aspect-Sentiment Analysis (JMASA) is a recently proposed task in the field of multimodal sentiment analysis (MSA), which aims to identify the aspect terms and their associated sentiment polarities expressed in the text-image pair~\cite{ju2021joint}. As shown in Fig.~\ref{fig:example}, we can extract two aspect-sentiment pairs from the given text-image pair, i.e., (\textit{Tana}, \textit{positive}) and (\textit{iPad Mini Giveaway}, \textit{positive}). 
Existing researches devote their efforts to two major obstacles when developing JMASA models: modality noise and semantic gap. 

For modality noise reduction, previous methods focus on two kinds of noise, i.e., instance- and feature-level noise. Some works calculate the relation score between the whole text and image, and then assign lower contribution weights to the images with lower similarity for multimodal fusion~\cite{ju2021joint}. This easily brings much redundant features, since each modality commonly contains limited informative clues for identifying aspects and sentiments~\cite{yu2022targeted}. For example in Fig.~\ref{fig:example}, it is sufficient to extract two aspect-sentiment pairs based on the visual regions with red boxes and underlined textual fragment. To this end, most recent works extract representative contents within each modality, such as salient visual objects~\cite{wu2020multimodal,ling-etal-2022-vision} and possible candidate aspects~\cite{zhou-etal-2023-aom}. However, their reliance on off-the-shelf tools, i.e., object detectors and NLP tools, not only limits their performance by the quality of extractors, but also inevitably filters out useful modality information. Therefore, we assume an ideal solution should be: \textit{using the complete inputs to model multimodal interaction while regulating pertinent information flow from unimodal inputs to multimodal features}. 

For text-image semantic gap reduction, either coarse- or fine-grained semantic alignment strategies have been widely studied. For the approaches simultaneously utilizing both of them~\cite{ju2021joint,yu2022targeted,ling-etal-2022-vision,zhou-etal-2023-aom}, they disentangle the aligning process into two steps: (1) first pre-train a coarse-grained alignment subtask by detecting whether two modalities are correlated or classifying sentiments for text-image pairs, (2) then apply various cross-attention techniques to generate word-aware visual representations and (visual) region-aware textual representations for aligning fine-grained inter-modal semantic. This separate pipeline requires additional labeled datasets for pre-training and hinders the practical application. Besides, cross attention lacks accuracy, since it inevitably attends to irrelevant/noisy contents that dominate cross-modal interaction~\cite{yang2022vision}. Therefore, it is necessary to explore an appropriate solution to \textit{jointly model coarse- and fine-grained semantic alignment without resort to cross attention}.

In this paper, we propose a novel framework for JMASA from an information-theoretic perspective, named \textbf{RNG}, which couples multi-level \textbf{n}oise \textbf{r}eduction with multi-grained semantic \textbf{g}ap \textbf{r}eduction. Specifically, to reduce multi-level modality noise, we (1) adopt a \textbf{Global Relevance Constraint} (GR-Con) based on similarity between two modalities to alleviate the noise from unrelated text-image pairs; and (2) design an \textbf{Information Bottleneck Constraint} (IB-Con) based on the Information Bottleneck (IB) principle to prevent intra-modal noisy features from flowing into multimodal features, since IB has been widely used to discard irrelevant information from input and preserve relevant information for prediction. Furthermore, we devise a \textbf{Semantic Consistency Constraint} (SC-Con) to capture inter-modal sentiment consistency via mutual information maximization. By means of coarse- and fine-grained contrastive learning strategies, RNG successfully models semantic alignment between a text and an image as well as textual words and visual regions, respectively. 

In summary, our main contributions are:
\begin{enumerate}
    \item We develop a fresh formulation via information theory for JMASA. To the best of our knowledge, we are the first to apply mutual information to simultaneously reduce intra-modal feature noise and inter-modal semantic gap in the field of fine-grained MSA.
    \item We propose a novel framework named RNG, which is built upon three carefully designed constraints: GR-Con, IB-Con, and SC-Con. They are responsible for instance-level noise reduction, feature-level noise reduction, and multi-grained semantic alignment, respectively.
    \item We validate the effectiveness of RNG on two benchmark datasets. The results demonstrate that our proposed method outperforms the state-of-the-art methods.
\end{enumerate}

\section{Related Work}
\subsection{Multimodal Aspect-based Sentiment Analysis}
Most conventional works of fine-grained sentiment analysis focus on analyzing opinionated texts~\cite{zhou2019span,liu2022mrce,liu2023him}. In recent years, users tend to express their opinions through a combination of texts and images. Undoubtedly, this trend has spurred the development of multimodal sentiment analysis~\cite{li2022amoa,li2023qap}. According to the characteristics of visual features utilized, previous methods can be roughly divided into two categories. (1) One line of works focus on using the entire image. For example, most methods employ the convolutional models, i.e., ResNet, to split each image into visual blocks on average~\cite{yu-etal-2020-improving-multimodal,ju2021joint,yang2022cross,zhou-etal-2023-aom}. Few recent methods utilize Transformer-based models, i.e., ViT, to equally divide each image into visual patches~\cite{yu2022dual}. Thus, each visual block or patch in an image can interact with the paired text. (2) The other line of works focus on using parts of an image. For instance, typical approaches adopt object detectors to extract salient visual objects with higher confidence, such as Faster R-CNN used by Ling et al.~\cite{ling-etal-2022-vision} and Mask R-CNN used by Wu et al.~\cite{wu2020multimodal}. In this way, only the retained visual regions can interact with text. 

\subsection{Mutual Information}
Mutual Information (MI) is a concept from information theory that aims to measure the relationship between two random variables $X$ and $Y$, denoting as $I(X; Y)$. Recently, various MI-based mechanisms have been explored. Tishby et al.~\cite{tishby2015deep} leverage the IB principle to explain the learning of deep neural networks by finding optimal internal representations between the input variable and output variable. Oord et al.~\cite{oord2018representation} design a contrastive learning loss termed InfoNCE for unsupervised representation learning, which is proven to be a lower bound of MI. 

\begin{figure*}[!t]
\centering
\includegraphics[scale=0.3]{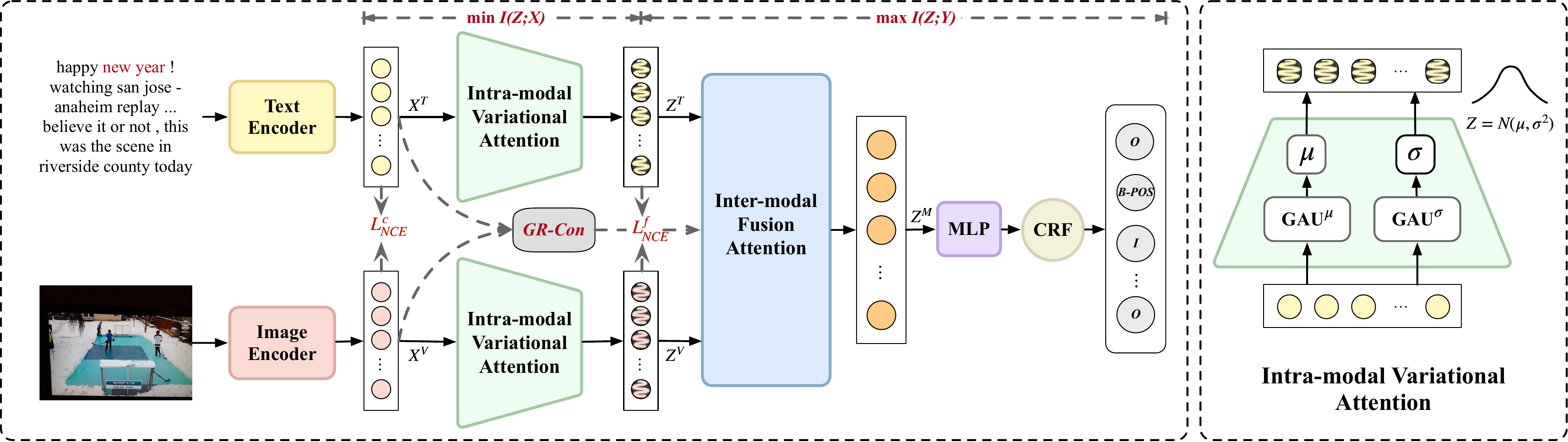}
\caption{An overview of our proposed RNG.}
\label{fig2:RNG}
\end{figure*}

\section{Methodology}

\subsection{Task Definition}
Given an input sample $X$ containing a text $T$ and an image $V$, JMASA aims to predict a label sequence $Y$ = $(y_1, y_2, ..., y_n)$, where $y_i \in \{$B-POS, B-NEU, B-NEG, I, O$\}$ and $n$ is the length of $T$. B, I, and O respectively denote the \textit{beginning}, \textit{inside}, and \textit{outside} of an aspect term, and POS, NEU, or NEG denotes the \textit{positive}, \textit{neutral}, or \textit{negative} sentiment polarity towards the aspect term. 

\subsection{Intuition Behind}
\label{sec:constraint}
To reduce multi-level modality noise and multi-grained semantic gap for developing a desirable JMASA model, we design following three constraints.

(1) \textbf{Global Relevance Constraint (GR-Con).} By calculating the similarity between the global text features and image features, GR-Con can measure the degree of an image being noisy. Thus, the text-image relevance will act as a global constraint to regulate multimodal fusion, which reduces instance-level noise from a holistic view. 

(2) \textbf{Information Bottleneck Constraint (IB-Con).} By applying the IB theory for intra-modal representation learning, IB-Con can naturally control information flow from raw intra-modal inputs to multimodal fusion, thus reducing feature-level noise. Under the constraint of IB-Con, the optimal latent representations $Z^T$ and $Z^V$ for the text and image can be generated by minimizing:
\begin{align}
    L_{IB}^T &= \beta_1 I(Z^T;X^T) - I(Z^T;Y), \\
    L_{IB}^V &= \beta_2 I(Z^V;X^V) - I(Z^V;Y).
\end{align}
Let $*$ stand for the textual or visual modality, i.e., $T$ or $V$. On the one hand, minimizing $I(Z^*;X^*)$ discourages $Z^*$ from learning noisy information from the input $X^*$. On the other hand, maximizing $I(Z^*;Y)$ encourages $Z^*$ to preserve sufficient task-related information for predicting the label $Y$. $\beta_1 \geq 0$ and $\beta_2 \geq 0$ control the trade-off. 

For simplicity, we assume $Z^T$ and $Z^V$ are two independent variables, thus $I(Z^X;Y)+I(Z^V;Y)$=$I(Z^T,Z^V;Y)$. The final objective function of IB-Con then can be written as,
\begin{align}
    L_{IB} &= L_{IB}^T + L_{IB}^V \\
           &= \beta_1 I(Z^T;X^T) + \beta_2 I(Z^V;X^V) - I(Z^T,Z^V;Y).
\end{align}

(3) \textbf{Semantic Consistency Constraint (SC-Con).} By maximizing MI between a matched text-image pair, i.e., $I(T;V)$, SC-Con can maximally ensure inter-modal semantic consistency between two modalities. Following the intuition from InfoNCE, we transform the maximization of MI into minimization of InfoNCE loss. Formally, the InfoNCE losses for text-to-image and image-to-text are defined as:
\begin{align}
    L_{nce}(T, V_{+}, \tilde{V}) &= -\mathbb{E}[\mathrm{log} \frac{\mathrm{exp}(\mathbf{\varepsilon}(T,V_{+})/\tau)}{\sum_{k=1}^{K_v} \mathrm{exp}(\mathbf{\varepsilon} (T,{\tilde{V}}_k)/\tau)}], \\
    L_{nce}(V, T_{+}, \tilde{T}) &= -\mathbb{E}[\mathrm{log} \frac{\mathrm{exp}(\mathbf{\varepsilon}(V,T_{+})/\tau)}{\sum_{k=1}^{K_t} \mathrm{exp}(\mathbf{\varepsilon} (V,{\tilde{T}}_k)/\tau)}],
\end{align}
where $V_{+}$/$T_{+}$ is a positive image/text example and $\tilde{V}$/$\tilde{T}$ is a set of negative image/text examples with the size of $K_v$/$K_t$. $\mathbf{\varepsilon}(\cdot,\cdot)$ denotes a similarity function and $\tau$ is a temperature hyperparameter. Overall, the total loss of InfoNCE is as follows, 
\begin{align}
    L_{NCE} &= \frac{1}{2} (L_{nce}(T, V_{+}, \tilde{V}) + L_{nce}(V, T_{+}, \tilde{T})).
\end{align}

Moreover, through defining the calculation of $\varepsilon(T, V)$ in a coarse- or fine-grained way, SC-Con can accordingly capture coarse- or fine-grained semantic consistency. Thus, the final objective function of SC-Con is $L_{SC}$ = $L_{NCE}^c + L_{NCE}^f$. 

\subsection{Model Specification}
\subsubsection{Modal Feature Extractor}
We take RoBERTa~\cite{liu2019roberta} and ViT~\cite{dosovitskiy2020image} as our text and image encoders. Firstly, the textual token sequence $T'$, with a length of $N$, is constructed by appending two special tokens $<$s$>$ and $<$/s$>$ to the beginning and end of a tokenized sequence. 
The visual patch sequence $V'$, with a length of $M$, is constructed by concatenating a special token [CLS] with the flattened visual patches. Then, the initial text representation $X^T\in \mathbb{R}^{N\times d}$ and image representation $X^V\in \mathbb{R}^{M\times d}$ are generated by: $X^T$ = $\mathrm{RoBERTa}(T')$ and $ X^V$ = $\mathrm{ViT}(V')$, on which a coarse-grained \textbf{SC-Con} can be imposed to reduce coarse-grained semantic gap.

\subsubsection{Intra-modal Variational Attention Layer}
To generate the optimal latent text/image representation $Z^T$/$Z^V$, we devise the variational attention encoder that makes the \textbf{IB-Con} trainable by the neural networks. Specifically, $Z^T$/$Z^V$ is sampled from a Gaussian distribution $\mathcal{N}(f^{\mu},f^{\sigma})$, where $f$ is the encoder network. 
Unlike previous MSA works that take Transformer-based networks as their modality encoders~\cite{yang2022cross}, we alternatively adopt Gated Attention Unit (GAU)~\cite{hua2022transformer} to implement $f$. For instance, taking $H$ as input, the output $H'$ of $f$ can be generated by $H'$ = $\mathrm{GAU}(H)$. The specific calculation process is as follows:
\begin{align}
     H' &= [(\phi_u(H W_u))\odot (A \phi_r(H W_r))]W_h, \\
     A &= \mathrm{softmax}(\mathcal{Q}(\phi_o(H W_o)){\mathcal{K}(\phi_o(H W_o))}^{\top}+b),
\end{align}
where $W_{u/r}\in \mathbb{R}^{d\times e}$, $W_h\in \mathbb{R}^{e\times d}$, and $W_o\in \mathbb{R}^{d\times l}$ are trainable parameters. $\odot$ is element-wise multiplication. $\phi$ denotes an element-wise activation function. Besides, $A\in \mathbb{R}^{N\times N}$ is the token-to-token attention weight matrix. $\mathcal{Q}$ and $\mathcal{K}$ apply per-dim scalars and offsets to $O$, and $b$ denotes the relative position bias. 

After that, the latent representation $Z^T$ of the textual IB-Con term is sampled from the Gaussian distribution $\mathcal{N}(\mu^T, [\mathrm{diag}(\sigma^T)]^2)$. Since the sampling procedure suffers from randomness, we use the re-parameterization trick~\cite{alemi2016deep,DBLP:journals/corr/KingmaW13} to allow sampling operation to be differentiable during back-propagation. Thus, we acquire $Z^T\in \mathbb{R}^{N\times d}$ by: 
\begin{align}
     z^T_i &= \mu^T_i + \sigma^T_i \odot \epsilon, \; \epsilon \sim \mathcal{N}(0, \mathrm{diag}(\mathbf{I})),
\end{align}
where the mean and standard deviation are learned by GAU, i.e., $\mu^T$=$\mathrm{GAU}^{\mu}(X^T)$ and $\sigma^T$=$\mathrm{GAU}^{\sigma}(X^T)$. 
Analogously, we can also obtain the visual representation $Z^V\in \mathbb{R}^{M\times d}$. 
To reduce fine-grained semantic gap, a fine-grained \textbf{SC-Con} is imposed on $Z^T$ and $Z^V$ before feeding them into inter-modal fusion attention layer. 

\subsubsection{Inter-modal Fusion Attention Layer}
To measure the degree of an image being noisy, we calculate the relevance score between the image and its paired text from a global view. 
Specifically, $x^T_1$ and $x^V_1$ are features of the token $<$s$>$ and the token [CLS], representing a text and an image from a holistic perspective. On the basis, we can calculate the global text-image correspondence $C$ by our devised score function, which is an inner product followed by a sigmoid function: $C$=$\,\mathrm{sigmoid}(x^T_1 \odot x^V_1)$.

Then, we impose our \textbf{GR-Con} on multimodal interaction with the correspondence $C$. To generate the multimodal representation, we adapt the aforementioned GAU into Cross-GAU, which can model inter-modal interactions between textual and visual modalities. Given the latent unimodal representations $Z^T$ and $Z^V$, Cross-GAU produces the multimodal representation $Z^M\in \mathbb{R}^{N\times d}$ by:
\begin{align}
     Z^M &= [(\phi_u(Z^T W_{u}^m)) \odot ((C\odot A') \phi_r(Z^V W_{r}^m))]W_z, \\
     A'&= \mathrm{softmax}(\mathcal{Q}(\phi_{t}(Z^T W_{t})){\mathcal{K}(\phi_{v}(Z^V W_{v}))}^{\top}+b),
\end{align}
where $W_{u/r}^m\in \mathbb{R}^{d\times d}$, $W_z\in \mathbb{R}^{d\times d}$, and $W_{t/v}\in \mathbb{R}^{d\times l}$ are trainable parameters. Besides, $A'\in \mathbb{R}^{N\times M}$ is the token-to-patch attention weight matrix.

\subsubsection{Output Layer}
To predict the label sequence $Y'$, we feed the multimodal representation $Z^M$ into a CRF~\cite{lafferty2001conditional} module as below:
\begin{align}
    \label{eq: crf}
    p(Y'|X) &= \frac{\mathrm{exp}(s(Z^M,Y'))}{\sum_{\hat{Y}\in \hat{Y}^A}\mathrm{exp}(s(Z^M, \hat{Y}))}, \\
    s(Z^M, Y) &= \sum_{i=0}^N G_{y_i,y_{i+1}} + \sum_{i=1}^N z^M_i\cdot W^{y_i},
\end{align}
where $G$ is a transition matrix and $\hat{Y}^A$ denotes all possible label sequences for the input sample. The trainable parameter $W^{y_i}\in \mathbb{R}^{d\times n_t}$ is used to compute the emission score from the token $t_i$ to the label $y_i$, where $n_t$ is the number of tags. 

\subsection{Training Objective}
On the top of our designed network, we construct the final training objective by integrating the losses used for task prediction and two MI-based constraints: $L$ = $L_{task} + L_{IB} + L_{SC}$.

\textbf{Task Prediction Loss.}
During training, we define the loss of task prediction by maximizing the log-probability of the correct label sequence: 
\begin{align}
    L_{task}=-\mathrm{log}p(Y'|X).
\end{align}

\textbf{Information Bottleneck Loss.}
For calculation of $L_{IB}$, since the computational challenge of mutual information, the IB objective is intractable to be directly optimized. Following the idea of Variational Information Bottleneck (VIB), we incorporate variational inference to approximate the information bottleneck constraint loss $L_{IB}$. 

Specifically, the minimization of the term $I(Z^T;X^T)$ is transformed to minimization of the variational upper bound that is a KL divergence term:
\begin{equation}
    \sum_{i=1}^N \mathrm{KL}(\mathcal{N}(\mu^T_i, [\mathrm{diag}(\sigma^T_i)]^2)||\mathcal{N}(0, \mathrm{diag}(\mathrm{\mathbf{I}}))).
\end{equation}
In the same way, we can also minimize the term $I(Z^V;X^V)$.

For the term $I(Z^T,Z^V;Y)$, maximizing it can be transformed to maximize the following variational lower bound:
\begin{align}
    \sum_{i=1}^N [\mathrm{log}(p(y_i|Z^T,Z^V)) + \sum_{\tilde{y}_i\in \tilde{Y}, \tilde{y}_i \neq y_i}\mathrm{log}(p(\tilde{y}_i|Z^T,Z^V))],
\end{align}
where $\tilde{Y}$ denotes the task label space. Here, we replace $X$ in Eq (\ref{eq: crf}) with the sampled $Z^T$ and $Z^V$ to calculate $p(\cdot|\cdot)$.

\textbf{Semantic Consistency Loss.}
Depending on the way to define similarity score function within $L_{SC}$, SC-Con can capture semantic consistency from multiple granularities. Take the definition of $\mathbf{\varepsilon}(T, V)$ as an instance, for coarse-grained semantic alignment, we follow CLIP~\cite{radford2021learning} and define $\mathbf{\varepsilon}(T, V)$ as follows:
\begin{align}
    \mathbf{\varepsilon}(T, V)=g_t(x^T_1)^{\top} g_v(x^{V}_1),
\end{align}
where $g_t(\cdot)$ and $g_v(\cdot)$ are two projectors. 

For fine-grained semantic alignment, we apply FILIP~\cite{yao2021filip} to define $\mathbf{\varepsilon}(T, V)$ by:
\begin{align}
    \mathbf{\varepsilon}(T, V) &= \frac{1}{N}\sum_{k=1}^N [z_i^T]_k^{\top}[z_j^V]_{m_k^T}, \\
    m_k^T &= \mathrm{argmax}_{0< r \leq M}[z_i^T]_k^{\top}[z_j^V]_r,
\end{align}
where $m_k^T$ indexes the visual patch that is the most similar to the $k$-th textual token.

\section{Experiment}

\subsection{Experimental Setups}
To validate the effectiveness of our proposed method, we conduct experiments on two public multimodal datasets, i.e., Twitter-2015 and Twitter-2017.\footnote{Twitter was renamed X in July 2023.} Table~\ref{tab:dataset} lists their detailed statistics. We employ Precision (P), Recall (R), and F1-score (F1) to evaluate the performance of our method.

\begin{table}[htbp]
\caption{Statistics of two multimodal Twitter datasets.}
\centering
\resizebox{7.2cm}{13.5mm}{
\begin{tabular}{ccccccccccc}
\toprule
        \multirow{2}{*}{\textbf {Dataset}} 
        & \multicolumn{3}{c}{\textbf{Twitter-2015}}  & \multicolumn{3}{c}{\textbf{Twitter-2017}}  \\ 
        \cmidrule(lr){2-4} \cmidrule(lr){5-7}  
        & Train  & Dev  & Test & Train  & Dev  & Test \\
        \midrule
        \#\,POS       & 928   & 303   & 317   & 1508  & 515  & 493   \\
        \#\,NEU       & 1883  & 670   & 607   & 1638  & 517  & 573   \\
        \#\,NEG       & 368   & 149   & 113   & 416   & 144  & 168   \\
        \#\,Sent      & 2101  & 674   & 674   & 1746  & 577  & 587   \\
        \bottomrule
\end{tabular}}
\label{tab:dataset}
\end{table}

We finetune all hyperparameters by grid searching. For the encoders RoBERTa and ViT, their learning rates are set to 3e-5 and 2.5e-5 respectively for Twitter-2015 and Twitter-2017. The learning rate of other trainable parameters on both datasets is set to 1e-4. We use the AdamW optimizer with the weight decay of 0.01 to train our model for 20 epochs. The maximum lengths of the token sequence and patch sequence are set to 60 and 197. The hidden size $d$ is 768 and the dimension $e$ is set to $2d$=1536. The dimension $l$ is 128. The temperature $\tau$ is set to 0.07.

\begin{table}
\caption{Experimental results for JMASA on two Twitter datasets ($\%$). 
        The results with $^\ddag$ and $^\dag$ are retrieved from \cite{yang2022cross} and \cite{ling-etal-2022-vision}.  
        For more fair comparison, we reproduce the methods with $^\P$ and report average results with 5 random seeds.}
    \centering
    \resizebox{8.8cm}{32mm}{
    \begin{tabular}{l|ccc|ccc}
    \toprule
    \multirow{2}*{\textbf{Model}} & \multicolumn{3}{c|}{\textbf{Twitter-2015}} & \multicolumn{3}{c}{\textbf{Twitter-2017}}  \\ 
    \cmidrule(lr){2-4} \cmidrule(lr){5-7} 
    & P & R & F1  & P & R & F1   \\
    \midrule
    \multicolumn{7}{c}{Text-only} \\
    \midrule
    D-GCN$^\ddag$~\cite{chen2020joint}           
        & 58.3  & 58.8  & 59.4   
        & 64.1  & 64.2  & 64.1   \\
    RoBERTa$^\ddag$~\cite{liu2019roberta}       
        & 61.8  & 65.3  & 63.5   
        & 65.5  & 66.9  & 66.2   \\
    \midrule
    \multicolumn{7}{c}{Multimodal} \\
    \midrule
    UMT-collapse$^\ddag$~\cite{yu-etal-2020-improving-multimodal}    
        & 60.4  & 61.6  & 61.0
        & 60.0  & 61.7  & 60.8   \\  
    OSCGA-collapse$^\ddag$~\cite{wu2020multimodal}  
        & 63.1  & 63.7  & 63.2
        & 63.5  & 63.5  & 63.5   \\ 
    JML$^\ddag$~\cite{ju2021joint}             
        & 65.4  & 64.0  & 64.7
        & 65.3  & 66.2  & 65.8   \\  
    CMMT$^\ddag$~\cite{yang2022cross}            
        & 64.6  & 68.7  & 66.5
        & 67.6  & 69.4  & 68.5   \\
    DTCA$^\P$~\cite{yu2022dual}         
        & 65.2  & 67.8  & 66.5
        & 68.3  & 69.4  & 68.9   \\
    VLP-MABSA$^\dag$~\cite{ling-etal-2022-vision}   
        & 65.1  & 68.3  & 66.6
        & 66.9  & 69.2  & 68.0   \\ 
    AoM$^\P$~\cite{zhou-etal-2023-aom}                
        & 66.7  & 69.0  & 67.8
        & 67.8  & 70.4  & 69.1   \\  
    \cmidrule(lr){1-7} 
    RNG                    
        & \textbf{67.8}  & \textbf{69.5}  & \textbf{68.6}
        & \textbf{69.5}  & \textbf{71.0}  & \textbf{70.2} \\   
    \bottomrule
    \end{tabular}}
    \label{tab:results-JMASA}
\end{table}

\subsection{Main Results}
We compare our method with two groups of state-of-the-art baseline methods, including \textbf{text-only methods} and \textbf{multimodal methods}. 
The average results across 5 runs with random initialization of our model are reported in Table~\ref{tab:results-JMASA}, from which we can observe that: 
(\romannumeral1) Our model consistently outperforms the baseline methods on both datasets, demonstrating the effectiveness of our proposed method. Compared with previous best results, RNG improves F1-score by 0.8\% and 1.1\% on Twitter-2015 and Twitter-2017, respectively.
(\romannumeral2) Multimodal methods gain remarkable advantages over text-only methods, highlighting the importance of visual information for understanding textual contents.
(\romannumeral3) RNG performing better than VLP-MABSA and AoM reveals two-fold superiority. The first one lies in reducing multi-level modality noise based on complete modality information, since VLP-MABSA and AoM only reduce feature-level noise based on extracted salient modality contents. The second one lies in reducing multi-grained semantic gap, because VLP-MABSA and AoM simply consider fine-grained cross-modal alignment.

\begin{table}[htbp]
\caption{Ablation study of each constraint in RNG ($\%$).}
    \centering
    \resizebox{8.1cm}{20mm}{
    \begin{tabular}{lcccccc}
    \toprule
    \multirow{2}*{\textbf{Model}} & \multicolumn{3}{c}{\textbf{Twitter-2015}} & \multicolumn{3}{c}{\textbf{Twitter-2017}} \\ 
    \cmidrule(lr){2-4} \cmidrule(lr){5-7} 
    & P & R & F1  & P & R & F1  \\
    \midrule
    \textbf{RNG}
        & 67.8  & 69.5  & 68.6
        & 69.5  & 71.0  & 70.2   \\  
    \midrule
    $\;$ w/o GR-Con
        & 66.7  & 69.7  & 68.0
        & 67.1  & 68.6  & 67.8   \\ 
    $\;$ w/o IB-Con
        & 65.2  & 67.0  & 66.1
        & 67.7  & 68.3  & 68.0    \\   
    $\;$ w/o SC-Con$^c$
        & 65.9  & 68.9  & 67.4
        & 67.9  & 70.1  & 69.0     \\
    $\;$ w/o SC-Con$^f$
        & 63.6  & 66.6  & 67.6
        & 68.9  & 69.7  & 69.3     \\
    $\;$ w/o SC-Con
        & 64.5  & 68.9  & 66.6
        & 69.4  & 70.1  & 68.7     \\
    $\;$ w/o all Con
        & 63.8  & 66.6  & 65.2
        & 67.5  & 67.6  & 67.6     \\
    \bottomrule
    \end{tabular}}
    \label{tab:abaltion-study}
\end{table}

\subsection{Ablation Study}
We study the efficacy of each constraint in RNG. As shown in Table~\ref{tab:abaltion-study}, RNG contributes significantly to the improvement of `w/o GR-Con' and `w/o IB-Con'. The performance gain is due to the instance- and feature-level noise reduction. Besides, imposing GR-Con on RNG is more beneficial to Twitter-2017, which contains more highly related text-image pairs. `w/o SC-Con$^c$', `w/o SC-Con$^f$', and `w/o SC-Con' degrade the model performance. This verifies that coarse- and fine-grained semantic alignment strategies are both crucial and complement to each other. Furthermore, the full model exhibits a 3\% F1-score discrepancy to `w/o all Con' on average. This sharp disparity suggests that integrating three constraints greatly promotes the extraction performance. 

\begin{figure}[htbp]
\centering
\begin{subfigure}{}
  \includegraphics[scale=0.5]{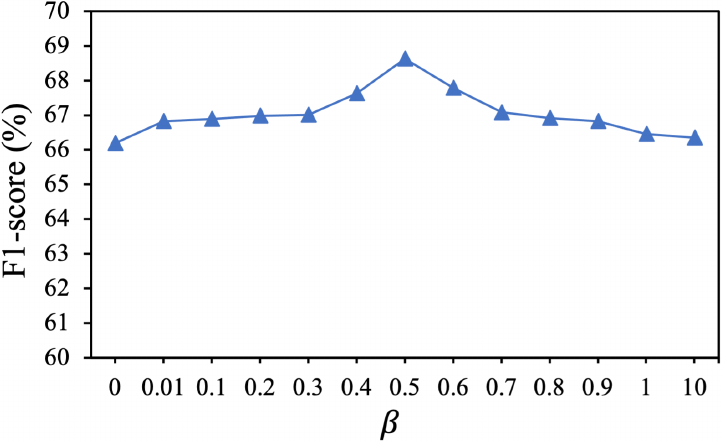}
  \centerline{(a) Twitter-15}
\end{subfigure}
\\
\vspace{0.5cm}
\begin{subfigure}{}
  \includegraphics[scale=0.5]{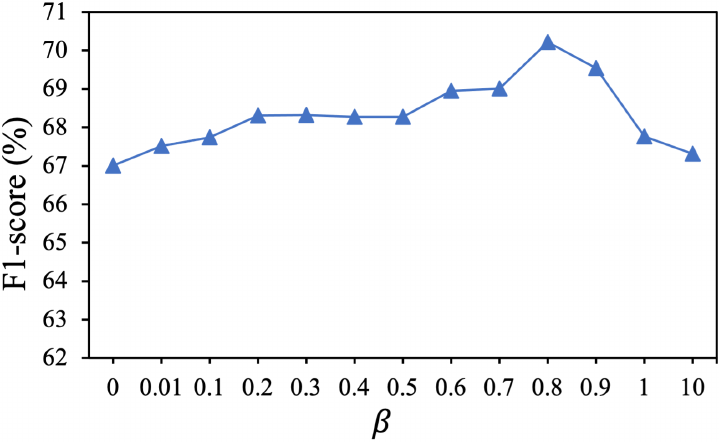}
  \centerline{(b) Twitter-17}
\end{subfigure}
\caption{F1-score against different $\beta$ on two datasets (\%).}
\label{fig:beta}
\end{figure}

\subsection{Hyperparameter Analysis}
We investigate the effect of the trade-off coefficients $\beta_1$ and $\beta_2$ in $L_{IB}$. For briefness, we set the same value $\beta$ for both of them. As shown in Fig.~\ref{fig:beta}, with $\beta$ increasing, F1 results of two datasets increase first. The optimal performances are attained when $\beta$ is set to 0.5 and 0.8 for Twitter-2015 and Twitter-2017, respectively. It is worth noting that the case of $\beta$=0 means omitting the minimization information constraint within IB-Con. The consequent performance decline confirms the significance of intra-modal feature denoising. In addition, the performance gradually drops as $\beta$ becomes larger, especially when $\beta$=10. The reason might be that overly compressing information from inputs potentially filters out some useful information for prediction.

\subsection{Case Study}
\begin{figure}[htbp]
\centering
\includegraphics[scale=0.38]{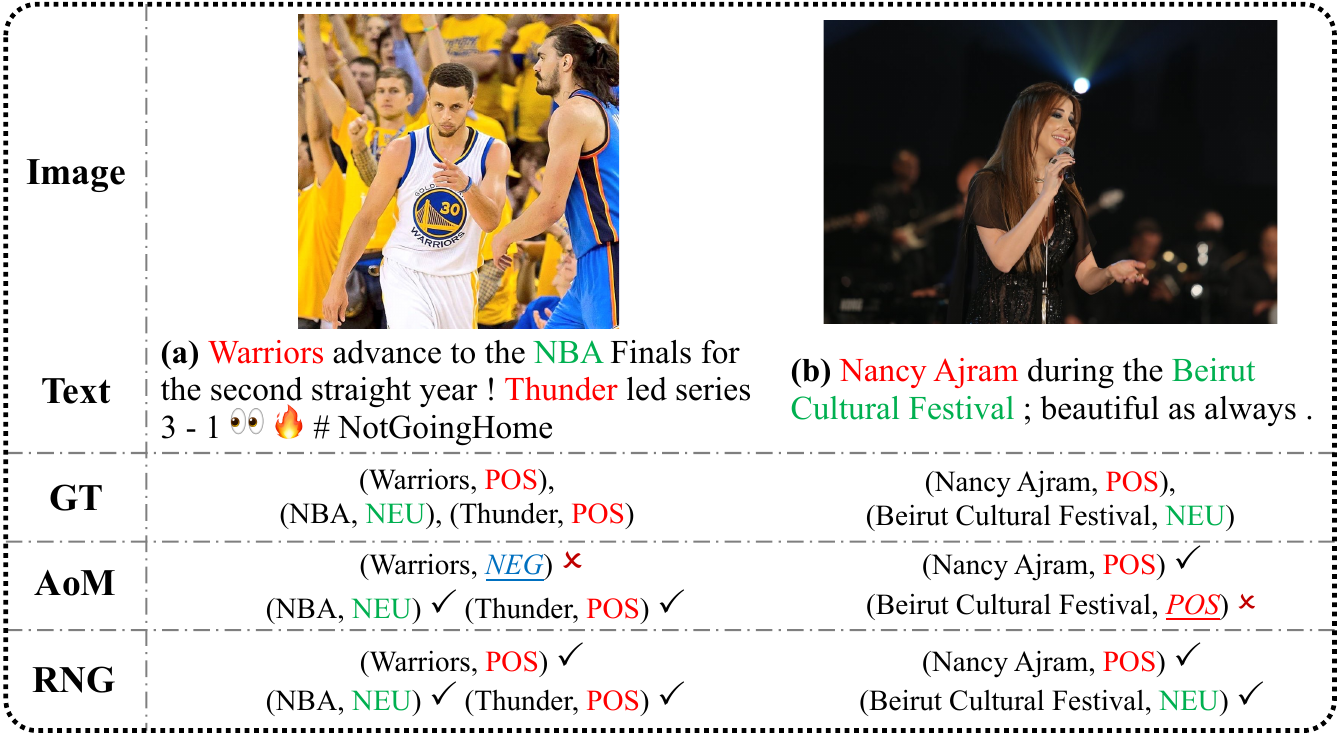}
\caption{Predictions of different methods on two test samples.}
\label{fig4:case_study}
\end{figure}

We display two samples with predictions from AoM and our RNG model in Fig.~\ref{fig4:case_study}. 
For case (a), AoM first extracts salient visual objects that are many NBA players. Then, the most salient visual object containing `Stephen Curry' interacts with `Warriors'. With the noise of an non-positive expression that
is conflicted with textual contents, sentiment classification towards `Warriors' suffers from negative influence.
For case (b), since AoM uses cross attention to align fine-grained semantic between retained visual objects and candidate aspects, the visual object containing `Nancy Ajram' with a positive expression probably dominates multimodal fusion, thus misleading the final prediction of `Beirut Cultural Festival'. 

\section{Conclusion}
This paper presents a multi-level modality noise and multi-grained semantic gap reduction architecture named RNG for JMASA. The RNG is built upon three well-designed constraints: Global Relevance Constraint (GR-Con), Information Bottleneck Constraint (IB-Con), and Semantic Consistency Constraint (SC-Con). These constraints regulate RNG to reduce instance-level modality noise, feature-level modality noise, and multi-grained semantic gap, respectively. Comprehensive experiments show that RNG outperforms the state-of-the-art methods.

\bibliographystyle{IEEEtran}
\bibliography{ICME}

\end{document}